\documentclass[conference]{IEEEtran}

\usepackage[utf8]{inputenc}
\usepackage{amsmath}
\usepackage{amsfonts}
\usepackage{amssymb}

\usepackage[T1]{fontenc}
\usepackage{textcomp}
\usepackage{gensymb}
\usepackage[normalem]{ulem}
\usepackage[hidelinks]{hyperref}

\usepackage[numbers]{natbib}

\usepackage{lipsum}

\usepackage{amsmath}
\usepackage[autostyle=true]{csquotes} 
\usepackage[pdftex]{graphicx}
\usepackage{xcolor}
\usepackage{color}
\usepackage{caption} 
\usepackage[parfill]{parskip}    
\usepackage{setspace} 
\usepackage{enumerate} 
\usepackage{booktabs} 
\usepackage{fancyhdr}
\usepackage{ulem}

\usepackage{float}
\usepackage{array}
\usepackage{tabularx}
\usepackage{diagbox}
\usepackage{multirow}
\usepackage{mathptmx}
\usepackage{mathrsfs}
\usepackage{mathtools}
\usepackage{amsmath}
\newcommand*{\Comb}[2]{{}^{#1}C_{#2}}%
\usepackage{subfig}
\usepackage{indentfirst}
\usepackage[super]{nth}
\usepackage{textcomp}
\usepackage{color,soul}

\usepackage{natbib}

\begin{document}

\twocolumn[
\begin{@twocolumnfalse}

\author{Reenul Reedha, Eric Dericquebourg, Raphael Canals, Adel Hafiane\\
\small{INSA CVL, University of Orleans, PRISME EA 4229, Bourges, 18022,France}}

\title{Vision Transformers For Weeds and Crops Classification Of High Resolution UAV Images}
\maketitle

\begin{abstract}
    Crop and weed monitoring is an important challenge for agriculture and food production nowadays. Thanks to recent advances in data acquisition and computation technologies, agriculture is evolving to a more smart and precision farming to meet with the high yield and high quality crop production. Classification and recognition in Unmanned Aerial Vehicles (UAV) images are important phases for crop monitoring. 
    Advances in deep learning models relying on Convolutional Neural Network (CNN) have achieved high performances in image classification in the agricultural domain. 
    Despite the success of this architecture, CNN still faces many challenges such as high computation cost, the need of large labelled datasets, ...  Natural language processing's transformer architecture can be an alternative approach to deal with CNN's limitations. Making use of the self-attention paradigm, Vision Transformer (ViT) models can achieve competitive or better results without applying any convolution operations. In this paper, we adopt the self-attention mechanism via the ViT models for plant classification of weeds and crops: red beet, off-type beet (green leaves), parsley and spinach. Our experiments show that with small set of labelled training data, ViT models perform better compared to state-of-the-art CNN-based models EfficientNet and ResNet, with a top accuracy of 99.8\% achieved by the ViT model.

\end{abstract}


\hrule
\end{@twocolumnfalse}]

\section{Introduction}
Agriculture is at the heart of scientific evolution and innovation to face major challenges for achieving high yield production while protecting plants growth and quality to meet the anticipated demands on the market \cite{Radoglou-Grammatikis2020}. However, a major problem arising in modern agriculture is the excessive use of chemicals to boost the production yield and to get rid of unwanted plants such as weeds from the field \cite{T2020}. Weeds are generally considered harmful to agricultural production \cite{Patel2016}. They compete directly with crop plants for water, nutrients and sunlight \cite{Iqbal}. Herbicides are often used in large quantities by spraying all over agricultural fields which has however shown various concerns like air, water and soil pollution and promoting weed resistance to such chemicals \cite{T2020}. If the rate of usage of herbicides remains the same, in the near future, weeds will become fully resistant to these products and eventually destroy the harvest \cite{Vrbnicanin2017}. This is why weed and crop control management is becoming an essential field of research nowadays \cite{Wang}.
\par
Automated crop monitoring system is a practical solution that can be beneficial both economically and environmentally. Such system can reduce labour costs by making use of robots to remove weeds and hence minimising the use of herbicides \cite{Wu2020}. The foremost step to an automatic weed control system is the detection and mapping of weeds on the field which can be a challenging part as weeds and crop plants often have similar colours, textures, and shapes \cite{Iqbal}. The use of Unmanned Aerial Vehicles (UAVs) has proved significant results for mapping weed density across a field by collecting RGB images (\cite{Donmez2021}, \cite{Bah2020}, \cite{Huang2018}, \cite{Huang2020}, \cite{Petrich2020}) or multispectral images (\cite{Puerto2020}, \cite{Ramirez2020}, \cite{Patidar2020}, \cite{Sa2017}, \cite{Sa2018}) covering the whole field. As UAVs fly over the field at an elevated altitude, the images captured cover a large ground surface area and these large images can be split into smaller tiles to facilitate their processing (\cite{DoSantosFerreira2017}, \cite{Milioto2017UAV}, \cite{Sivakumar2020}) before feeding them to learning algorithms to identify and classify a weed from a crop plant. 
\par

In the agricultural domain, the main approach to plant detection is to first extract vegetation from the image background using segmentation and then distinguish crops from the weeds \cite{Kang2021}. Common segmentation approaches use multispectral information to separate the vegetation from the background (soil and residuals) \cite{Kerkech2019}. However, weeds and crops are difficult to distinguish from one another even while using spectral information because of their strong similarities \cite{Hamuda2016}. This point has also been highlighted in \cite{Wang}, in which the authors reported the importance of using both spectral and spatial features to identify weeds in crops. Recently, deep learning (DL) became an essential approach in image classification, object detection and recognition \cite{LeCun2015}, \cite{Hochreiter1997} notably in the agricultural domain \cite{Hasan2021}. DL models with CNN-like architectures, ruling in computer vision tasks so far, have been the standard in image classification and object detection \cite{Lecun}, \cite{Krizhevsky}, \cite{Kaiming}. CNN uses convolutional filters on an image to extract important features to understand the object of interest in an image with the help of convolutional operations covering key properties such as local connection, parameters (weight) sharing and translation equivariance \cite{Lecun89}, \cite{LeCun2015}. Most of the papers covering weed detection or classification make use of CNN-based model structures \cite{Nkemelu}, \cite{Suh2018}, \cite{DianBah2018} such as AlexNet \cite{Krizhevsky}, VGG-19, VGG-16 \cite{Simonyan}, GoogLeNet \cite{Christian_Googlenet}, ResNet-50, ResNet-101 \cite{Kaiming} and Inception-v3 \cite{Christian}. 
\par

On the other hand, attention mechanism has seen a rapid development particularly in natural language processing (NLP) \cite{Hu2019} and has shown spectacular performance gains when compared to previous generation of state-of-the-art models \cite{Vaswani2017}. In vision applications the use of attention mechanism has been much more limited, due to the high computational cost as the number of pixels in an image is much larger than the number of units of words in NLP applications. This makes it impossible to apply standard attention models to images. A recent survey of applications of transformer networks in computer vision can be found in \cite{Khan2021}. The recently-proposed vision transformer (ViT) appears to be a major step towards adopting transformer-attention models for computer vision tasks \cite{Dosovitskiy2020}. Considering image patches as units of information for training is a groundbreaking method compared to CNN-based models considering image pixels. ViT incorporates image patches into a shared space and learns the relation between these patches using self-attention modules. Given massive amounts of training data and computational resources, ViT was shown to surpass CNNs in image classification accuracy \cite{Dosovitskiy2020}. Vision transformer models have not been used yet for weeds and crops classification of high resolution UAV images.  

\par
In this paper, we adopt the self-attention paradigm via vision transformers for the classification of images of weeds and different crops: red leaves beet, green leaves beet, parsley and spinach, taken by a high resolution digital camera mounted on a UAV. We make use of the self-attention mechanism on small, labelled plant images using the convolutional-free ViT model, showing its outperformance compared to current state-of-the-art CNN-based models ResNet and EfficientNet. Furthermore, we show that when training with fewer number of labelled images, the ViT model performs better than the CNN-based models. The rest of the paper is organised as follows. Section 2 presents the materials and methods used as well as a brief description of the self-attention mechanism and the vision transformer model architecture. The experimental results and analysis are presented in Section 3. Section 4 reflects the conclusion and work perspectives. 
    
\section{Materials and Methods}
This part outlines the acquisition, preparation and manual labelling of the dataset acquired from a high resolution camera mounted on a UAV. It also presents a brief description of the self-attention paradigm and the vision transformer model architecture. 

    \subsection{Image collection and annotation}
    The study area is located at the agricultural fields of beet, parsley and spinach present in the region of Centre Val de Loire, in France. It is also a region with many pedo-climatic advantages: the region has limited rainfall and clay-limestone soils with good filtering capacity. Irrigation is also offered on 95\% of the plots.
    
\begin{figure}[H]
    \centering
    \subfloat[Starfury Drone]{{\includegraphics[width=3cm]{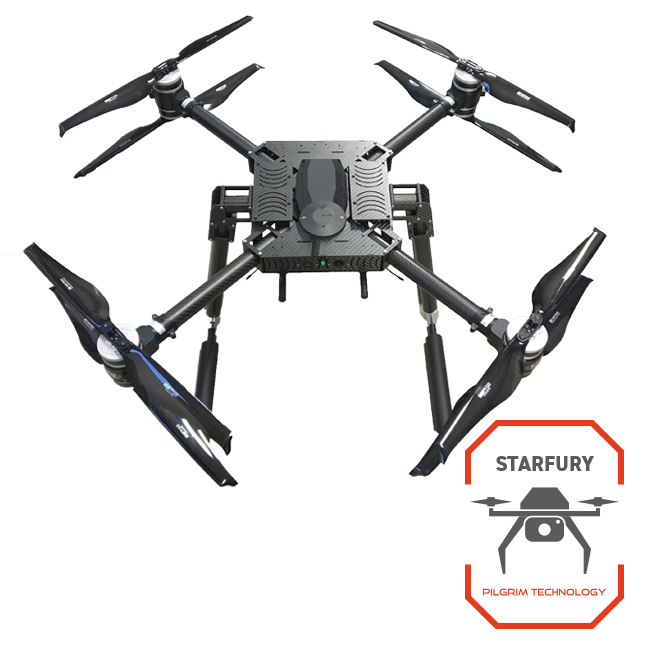} }}
    \qquad
    \subfloat[Sony ILCE-7R Camera]{{\includegraphics[width=3cm]{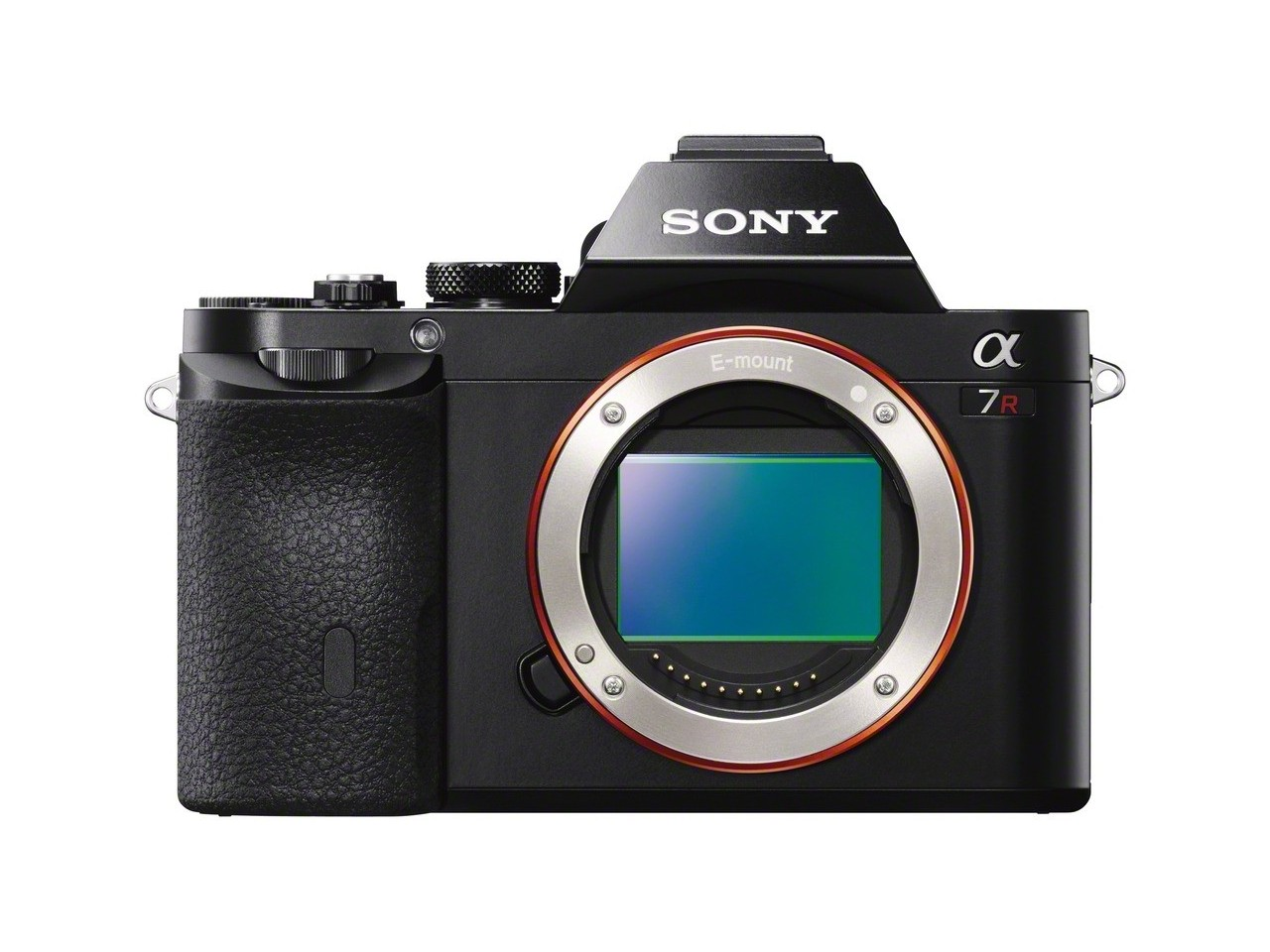} }}
    \caption{Apparatus used for data acquisition.}
    \label{fig:Apparatus used for data acquisition}
    \end{figure}

\begin{figure}[H]
        \centering
        \includegraphics[width=0.5\textwidth]{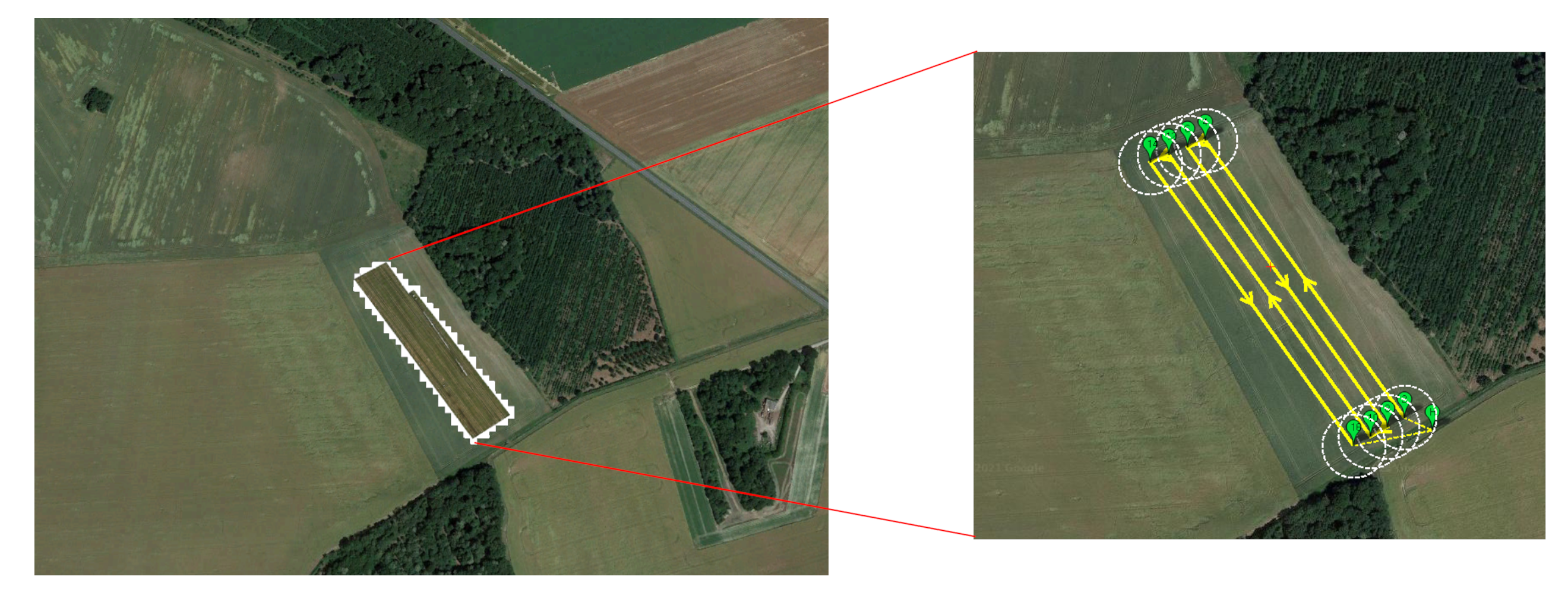}
        \caption{Overlay of the orthophoto on google earth of the spinach plot (left) and the flight plan (right) across a spinach field (the images are taken along the yellow lines at regular intervals to ensure sufficient overlapping).}
        \label{fig:Spinach field routing}
    \end{figure}

    To survey the study areas, a "\textit{Starfury}", Pilgrim technologies UAV was equipped with a \textit{Sony ILCE-7R}, 36 mega pixels camera as shown in \autoref{fig:Apparatus used for data acquisition}. The camera is mounted on a 3-axis stabilised brushless gimbal on the drone in order to keep the camera axis stable even during strong winds. The drone was flown at an altitude of 30 m over the beet field in a light morning fog and at an altitude of 20 m in a sunny weather for the parsley and spinach fields. The drone followed a specific flight plan and the camera captured RGB images at regular intervals as shown in \autoref{fig:Spinach field routing}. The images captured have respectively a minimum longitudinal and lateral overlapping of 70\% and 50-60\% depending on the fields vegetation coverage and homogeneity, assuring a better and complete coverage of the whole field of 4ha (40 000 m²) and improving the accuracy of the field orthophoto generation. 
    
    \begin{figure}[H]
        \centering
        \includegraphics[width=0.5\textwidth]{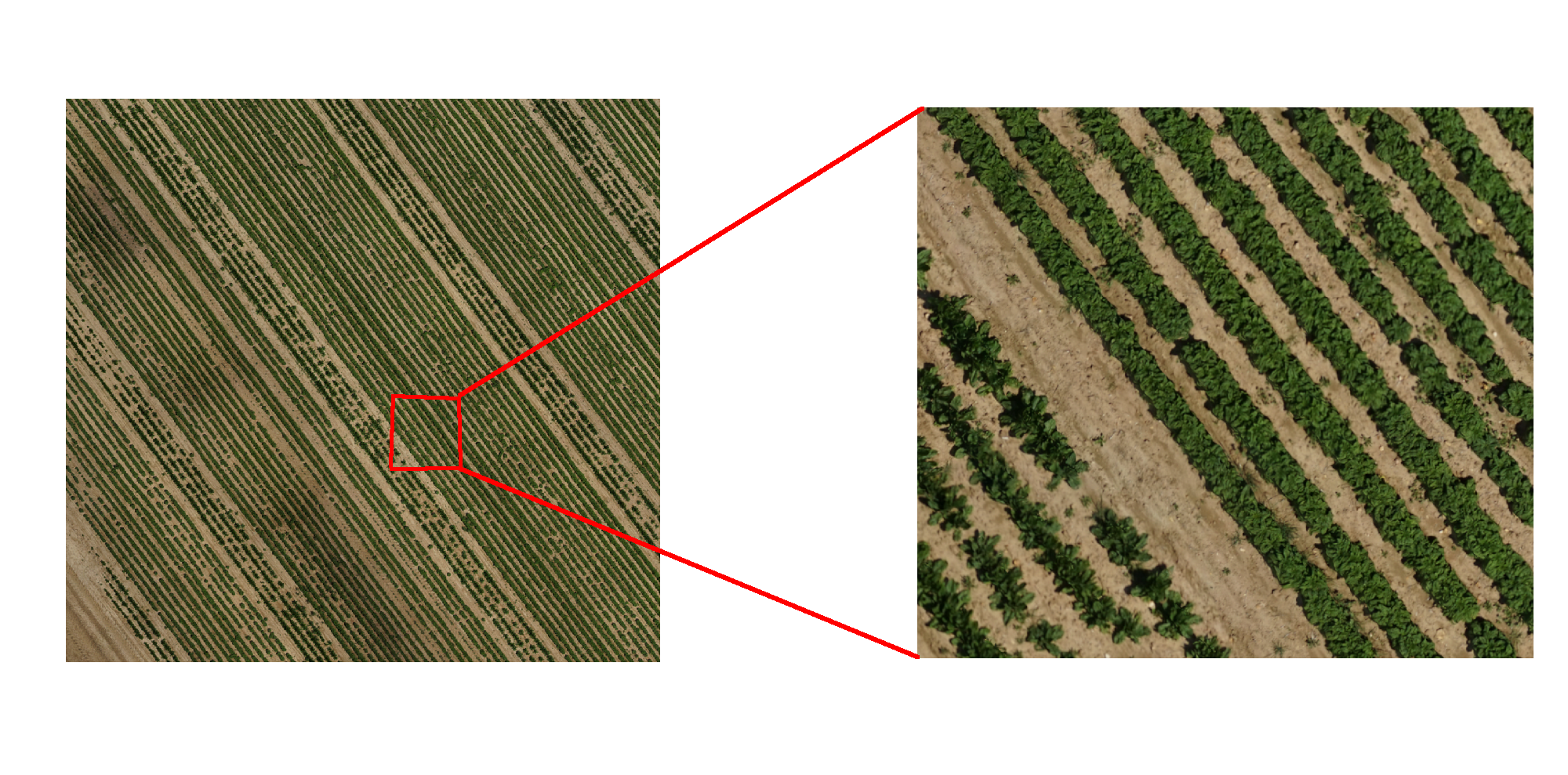}
        \caption{Example of image captured from a spinach study site.}
        \label{fig:Beet field}
    \end{figure}
    
    The data was then manually labelled using bounding boxes with the tool LabelImg (\url{https://github.com/tzutalin/labelImg}) and then divided into 5 classes as shown in \autoref{fig:Samples of each class} and \autoref{tab:Class Distribution} below. 
    
    \begin{figure}[H]
    \begin{tabular}{cc}
    \includegraphics[width=0.2\textwidth]{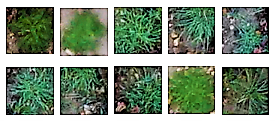} &   \includegraphics[width=0.2\textwidth]{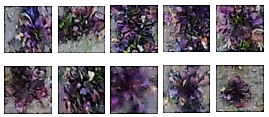} \\
    \small{(a) Weeds} & \small{(b) Beet} \\[6pt]

    \includegraphics[width=0.2\textwidth]{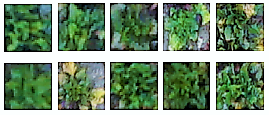} &   \includegraphics[width=0.2\textwidth]{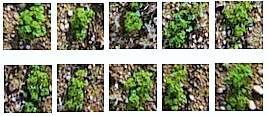} \\
    \small{(c) Off-type green leaves beet} & \small{(d) Parsley} \\[6pt]

    \multicolumn{2}{c}{\includegraphics[width=0.2\textwidth]{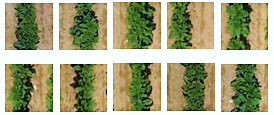} }\\
    \multicolumn{2}{c}{\small{(e) Spinach}}

    \end{tabular}
    \caption{This overview shows sample images patches of all 5 classes of our custom dataset. The images measure 64 x 64 pixels. Each class contains 3200 to 4000 images.}
    \label{fig:Samples of each class}
    \end{figure}
    
    \begin{table}[H]
    \centering
    \begin{tabularx}{0.4\textwidth}{ 
  | >{\raggedright\arraybackslash}X 
  | >{\centering\arraybackslash}X | }
    \hline
    \textbf{Class} & \textbf{Number} \\ \hline
    Weed & 4 000 \\ \hline
    Beet & 4 000 \\ \hline
    Off-type Beet & 3 265 \\ \hline
    Parsley & 4 000 \\ \hline
    Spinach & 4 000 \\ \hline
    \end{tabularx}
    \caption{Class Distribution}
    \label{tab:Class Distribution}
    \end{table}
    
    The labelled dataset has been classified in 5 folders for each class label and each containing an equal number of images as presented in \autoref{tab:Class Distribution}. We have 16.9\% off-type beet plants (obtained by data augmentation - flips and rotations from \textit{765} labelled data) and equally 20.8\% images for the four other classes. For a total of 19 265 images of size 64x64. The dataset is then divided into training, validation and testing sets as shown in \autoref{fig:train_val_test}.  
    
    \subsection{Image preprocessing}
    Due to the huge labour cost of the whole process of supervision training, artificial data augmentation was used in the experiment to generate additional new images and increase the amount of data to solve the problem of insufficient agricultural image dataset of weeds and crops. Image data augmentation is not only used to expand the training dataset and attempting to cover real-world scenarios but also to improve the performance and ability of the model to generalise on agricultural images as they can vary a lot depending on the soil, environment, seasons and climate conditions.

    As a data preprocessing method, data augmentation plays an important role in deep learning \cite{Pan2010}. In general, effective data augmentation can better improve the robustness of the model and obtain stronger generalisation capabilities. We thus employed data augmentation strategies, which have been widely used in practice, so as to enrich the datasets. After normalising each image, the following steps have been applied: random rotation, random resized crop, random horizontal flip, colour jitters and rand augment (\citeauthor{Cubuk2020}). This technique is implemented using \textit{keras ImageDataGenerator}, generating augmented images on the fly while the model is still in the training stage.

    \subsection{Self attention for weeds detection}
    Attention mechanism is becoming a key concept in the deep learning field \cite{Niu2021}. Attention was inspired by the human perception process where the human tends to focus on parts of information, ignoring other perceptible parts of information at the same time. The attention mechanism has had a profound impact on the field of natural language processing, where the goal was to focus on a subset of important words. The self-attention paradigm has emerged from the concepts attention showing improvement in the performance of deep networks \cite{Vaswani2017}. 
    
    Lets denote a sequence of n entities ($x_1, x_2,..., x_n$) by $X \in \mathbb{R}^{n\times d}$, where d is the embedding dimension to represent each entity. The goal of self-attention is to capture the interaction amongst all n entities by encoding each entity in terms of the global contextual information. This is done by defining three learnable weight matrices, Queries ($W^Q \in \mathbb{R}^{n \times d_q}$), Keys ($W^K \in \mathbb{R}^{n\times{d_k}}$) and Values ($W^V \in \mathbb{R}^{n\times{d_v}}$). The input sequence X is first projected onto these weight matrices to get $Q = XW^Q, K = XW^K and V = XW^V$.
    
    The attention matrix $A \in \mathbb{R}^{n\times{d_v}}$ indicates a score between N queries Q and $K^T$ keys representing which part of the input sequence to focus on.
    
    \begin{equation} \label{eq:1}
        A(Q, K) = \sigma(QK^T) 
    \end{equation}
    
    where $\sigma$ is an activation function, usually $softmax()$. To capture the relations among the input sequence, the values V are weighted by the scores from Equation \ref{eq:1}. Resulting in \cite{Dosovitskiy2020}, 
    
    \begin{equation} \label{eq:2}
    \begin{split}
        SelfAttention(Q, K, V) = A(Q, K) \cdot V 
        \\
        \Rightarrow SelfAttention(Q, k, V) = softmax(\frac{QK^T}{\sqrt{d_k}})\cdot V
    \end{split}
    \end{equation}
    \par
    where $d_k$ is dimension of the input queries. 
    
    If each pixel in a feature map is regarded as a random variable and the paring covariances are calculated, the value of each predicted pixel can be enhanced or weakened based on its similarity to other pixels in the image. The mechanism of employing similar pixels in training and prediction and ignoring dissimilar pixels is called the self-attention mechanism. It helps to relate different positions of a single sequence of image patches in order to gain a more vivid representation of the whole image \cite{ChengDL16}. 
    
    The transformer network is an extension of the attention mechanism from Equation \ref{eq:2} based on the Multi-Head Attention operation. It is based on running k self-attention operations, called “heads”, in parallel, and project their concatenated outputs \cite{Vaswani2017}. This helps the transformer jointly attend to different information derived from each head. The output matrix is obtained by the concatenation of each attention heads and a dot product with the weight $W^O$. Hence, generating the output of the multi-headed attention layer. The overall operation is summarised by the equations below \cite{Vaswani2017}. 
    
    \begin{equation}
    \begin{split}
        MultiHead(Q, K, V) = Concat(head_1, ..., head_h)W^O
        \\
        where\space head_i = Attention(QW^Q_i, KW^K_i, VW^V_i)
    \end{split}
    \end{equation}
    where $W^Q_i, W^K_i, W^V_i$ are weight matrices for queries, keys and values respectively and $W^O \in \mathbb{R}^{hd_v \times d_{model}}$.
    
    By using the self-attention mechanism, global reference can be realised during the training and prediction of models. This helps in reducing by a considerable amount training time of the model to achieve high accuracy \cite{Dosovitskiy2020}. The self-attention mechanism is an integral component of transformers, which explicitly models the interactions between all entities of a sequence for structured prediction tasks. Basically, a self-attention layer updates each component of a sequence by aggregating global information from the complete input sequence. While, the convolution layers' receptive field is a fixed $K\times K$ neighbourhood grid, the self-attention's receptive field is the full image. The self-attention mechanism increases the receptive field compared to the CNN without adding computational cost associated with very large kernel sizes \cite{Zhang2019}. Furthermore, self-attention is invariant to permutations and changes in the number of input points. As a result, it can easily operate on irregular inputs as opposed to standard convolution that requires grid structures \cite{Khan2021}. 
    
    \begin{figure}[H]
        \centering
        \includegraphics[width=0.3\textwidth]{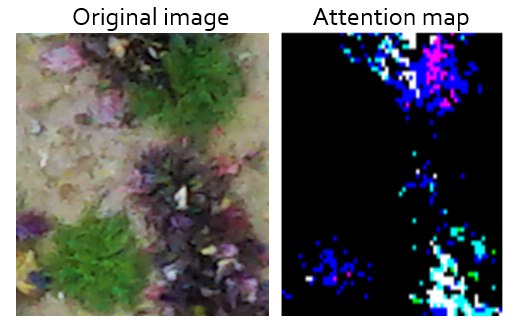}
        \caption{Attention mechanism on an image patch (left) containing weeds (in green) and beet plant (in red). With the original image on the left and the attention map on the right obtained with ViT-B16 model. The attention map shows the model's attention on the different plants: with a dark blue and purple colour pixel representing the attention on the weeds and a light blue colour pixel representing the beet plant.}
        \label{fig:Attention map on an overexposed weed patch}
    \end{figure}
    
    
    \begin{figure}[H]
        \centering
        \includegraphics[width=0.4\textwidth]{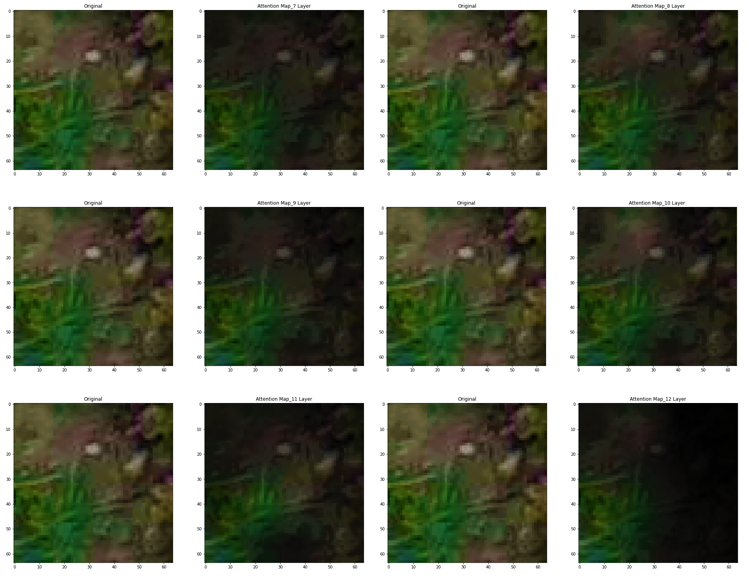}
        \caption{Attention map generated from layers 7 to 12 of the ViT-B16 model on an image of a weed.}
        \label{fig:Attention map generated from layers 7 to 12 of the ViT-B16 model}
    \end{figure}
    
    Average attention weights of all heads mean heads across layers and the head in the same layer. Basically, the area has every attention in the transformer which is called attention pattern or attention matrix. When the patch of the weed image is passed through the transformer, it will generate the attention weight matrix for the image patches. For example, when patch 1 is passed through the transformer, self-attention will calculate how much attention should pay to others (patch 2, patch 3, ...). And every head will have one attention pattern as shown in \autoref{fig:Attention map generated from layers 7 to 12 of the ViT-B16 model} and finally, they will sum up all attention patterns (all heads). We can observe that the model tries to identify the object (weed) on the image and tries to focus its attention on it (as it stands out from the background).
    
    An attention mechanism is applied to selectively give more importance to some of the locations of the image compared to others, for generating caption(s) corresponding to the image. And consequently, this helps to focus on the main differences between weeds and crops in an images and improves the learning of the model to identify the contrasts between these plants. This mechanism also helps the model to learn features faster, and eventually decreases the training cost \cite{Dosovitskiy2020}. 
    
    \subsection{Vision Transformers}
    
     \begin{figure*}
        \centering
        \includegraphics[width=0.65\textwidth]{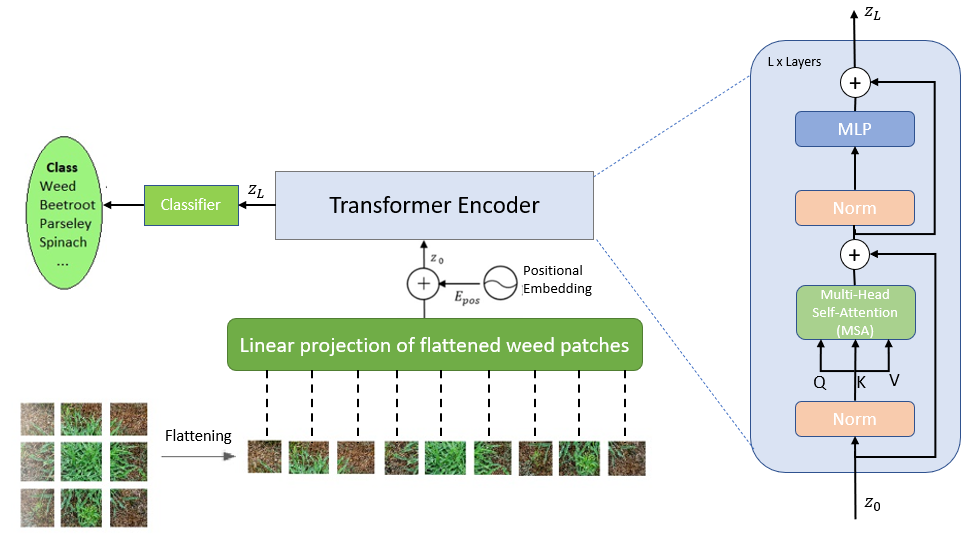}
        \caption{ViT model architecture based on original ViT model \cite{Dosovitskiy2020}.}
        \label{fig:ViT model architecture}
    \end{figure*}
    
     \begin{figure*}
        \centering
        \includegraphics[width=0.45\textwidth]{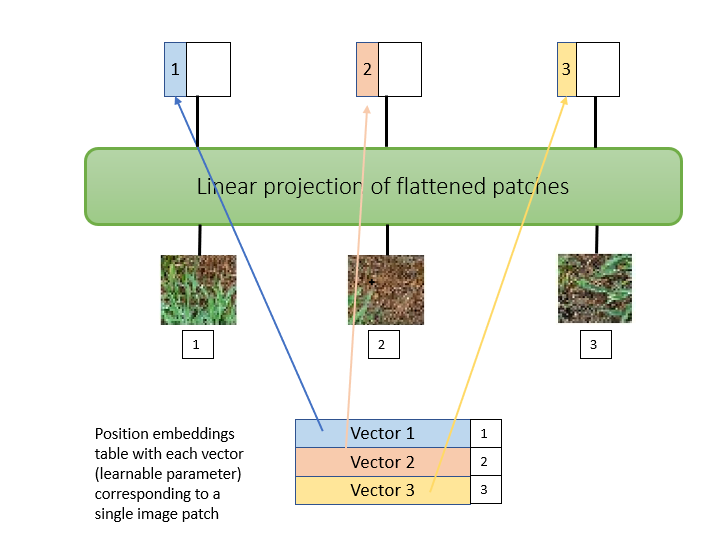}
        \caption{Positional embeddings as vector representations.}
        \label{fig:Positional embeddings as vector representations}
    \end{figure*}
    
    Transformer models were major headway in natural language processing (NLP). They became the standard for modern NLP tasks and they brought spectacular performance yields when compared to the previous generation of state-of-the-art models \cite{Vaswani2017}. Recently, it was reviewed and introduced to computer vision and image classification aiming to show that this reliance on CNNs is not necessary anymore in object detection or image classification and a pure transformer applied directly to sequences of image patches can perform very well on image classification tasks \cite{Dosovitskiy2020}.
    
    \autoref{fig:ViT model architecture} presents the architecture of the vision transformer used in this paper for weed and crop classification. It is based on the first developed ViT model by \citet{Dosovitskiy2020}. The model architecture consists of 7 main steps. Firstly, the input image is split into smaller fixed-size patches. Then each patch is flattened into a 1-D vector. The input sequence consists of the flattened vector (2D to 1D) of pixel values from a patch of size 16×16. 
    \par
    For an input image,
    \begin{equation}
        (x) \in \mathbb{R}^{H \times W \times C}
    \end{equation}
    
    and patch size $P$, N image patches are created 
    \begin{equation}
        (x)_P \in \mathbb{R}^{N \times P \times P \times C}
    \end{equation}
     
    with
    \begin{equation}
        N = \frac{HW}{P \times P}
    \end{equation}
    where $N$ is the sequence length (token) similar to the words of a sentence, $(H,W)$ is the resolution of the original image and $C$ is the number of channels \cite{Dosovitskiy2020}.  
    \par
    Afterwards, each flattened element is then fed into a linear projection layer that will produce what is called the “patch embedding”. There is one single matrix, represented as ‘E’ (embedding) used for the linear projection. A single patch is taken and first unrolled into a linear vector as shown in \autoref{fig:Positional embeddings as vector representations}. This vector is then multiplied with the embedding matrix E. The final result is then fed to the transformer, along with the positional embedding. In the \nth{4} phase, the position embeddings are linearly added to the sequence of image patches so that the images can retain their positional information. It injects information about the relative or absolute position of the image patches in the sequence. The next step is to attach an extra learnable (class) embedding to the sequence according to the position of the image patch. This class embedding is used to predict the class of the input image after being updated by self-attention. Finally, the classification is performed by just stacking a multilayer perceptron (MLP) head on top of the transformer, at the position of the extra learnable embedding that has been added to the sequence.

\section{Performance evaluation}
We made use of recent implementations of ViT-B32 and ViT-B16 models as well as EfficientNet and ResNet models. The algorithms were built on top of a Tensorflow 2.4.1 and Keras 2.4.3 frameworks using Python 3.6.9. To run and evaluate our methods, we used the following hardware; an Intel Xeon(R) CPU E5-1620 v4 3.50 GHz x 8 processor (CPU) with 16 GB of RAM, and a graphics processing unit (GPU) NVIDIA Quadro M2000 with an internal RAM of 4 GB under the Linux operating system Ubuntu 18.04 LTS (64 bits). 

All models were trained using the same parameters in order to have an unbiased and reliable comparison between their performance. The initial learning rate was set to 0.0001 with a reducing factor of 0.2. The batch size was set to 8 and the models were trained for 100 epochs with an early stopping after a wait of 10 epochs without better scores. The models used, ViT-B16, ViT-B32, EfficientNet B0, EfficientNet B1 and ResNet 50 were loaded from the keras library with pre-trained weights of "ImageNet". 

    \subsection{Cross Validation}
    The experiments have been carried out using the cross-validation technique to ensure the integrity and accuracy of the models. We used stratified K-Fold with the variation of the proportion of training and validation folders. Therefore, the dataset was divided into k folders using certain percentage of the data as a validation set and the rest as a training set. Stratified is to ensure that each fold contains the same proportion of each label. Thus stratified K-Fold is one of the best approaches in ensuring that the data is well balanced and shuffled in each folds before splitting them into validation and training sets. 

    \begin{figure}[H]
        \centering
        \includegraphics[width=0.5\textwidth, height=4cm]{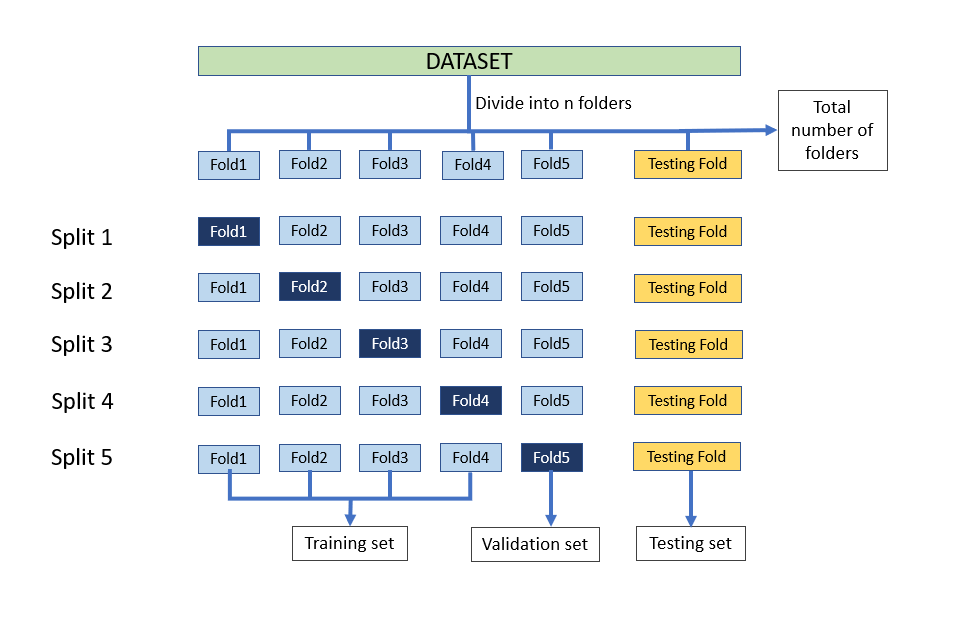}
        \caption{Stratified five-folds cross validation, leaving one out for validation and the remaining 4 folds are used for training. Dark blue representing validation folds, light blue colour folds are used as training set and yellow colour folds are used as testing set containing unprocessed images. This generates 5 trained models.}
        \label{fig:Stratified K-Fold Cross Validation leave one out}
    \end{figure}

\begin{figure}[H]
        \centering
        \includegraphics[width=0.5\textwidth]{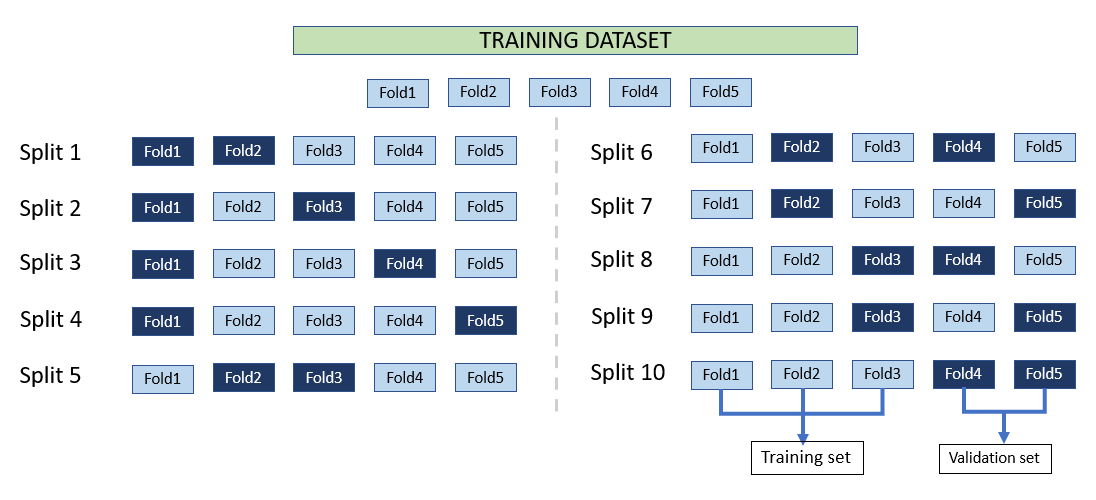}
        \caption{Stratified five-folds cross validation and leaving two out as validation set and the rest are used for training resulting in 10 different models. Dark blue representing validation folds and light blue colour folds are used as training set.}
        \label{fig:Stratified K-Fold Cross Validation leave two out}
    \end{figure}

\begin{figure}[H]
        \centering
        \includegraphics[width=0.4\textwidth]{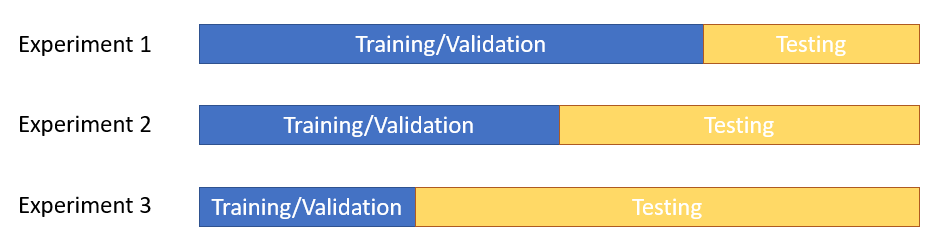}
        \caption{Variation of training/validation set and testing image set for 3 sets of experiments. The training/validation set is used for the cross validation as shown in \autoref{fig:Stratified K-Fold Cross Validation leave one out} and \autoref{fig:Stratified K-Fold Cross Validation leave two out}.}
        \label{fig:train_val_test}
    \end{figure}

    The data was first shuffled randomly and divided equally into 5 folders, each containing an equal number of classes and the performances of the tested models were evaluated using the stratified five-folds cross validation leaving $k$ folds as validation set (where $1 \leq k \leq 4$). \autoref{fig:Stratified K-Fold Cross Validation leave two out} shows how the data was splitted and divided into 5 folders each containing and equal number of classes. Using \autoref{combination} (with $n=5$ and $k=2$), this results in 10 training models. Increasing the value of $k$ (number of validation folders), decreases the number of training folders and thus forces the model to train on a smaller dataset. This helps to evaluate how well the models perform on reduced training datasets and their capacities to extract features from a few images. The number of combinations (splits) of the train-validation is as follows:  

    \begin{equation} \label{combination}
        \large \Comb{n}{k} = \frac{n!}{k!(n-k)!}
    \end{equation}
    where $n$ is the number of folders and $k$ is the number of validation folds.

  The labelled dataset is divided into training, validation and testing sets for three sets of experiments. The number of testing images is increased for each experiments, consequently decreasing the number of training and validation images. Experiment 1 contains 12 844 training images, Experiment 2 contains half the dataset for testing (9633 images) and 7706 training images and Experiment 3 contains only 4535 training images for 13 596 testing images. Each set of experiments is then trained using the cross validation technique leaving one fold as validation set.
    
    \subsection{Evaluation metrics}
    In the collected dataset, each image has been manually classified into one of the categories: weeds, off-type beet (green leaves beet), beet (red leaves), parsley or spinach, called ground-truth data. By running the classifiers on a test set, we obtained a label for each testing image, resulting in the predicted classes. The classification performance is measured by evaluating the relevance between the ground-truth labels and the predicted ones resulting in classification probabilities of true positives (TP), false positives (FP) and false negatives (FN). We then calculate a recall measure representing how well a model correctly predicts all the ground-truth classes and a precision representing the ratio of how many of the positive predictions were correct relative to all the positive predictions. 
    
    \begin{equation}
        \begin{split}
            Precision = \frac{TP}{TP+FP}
            \\
            Recall = \frac{TP}{TP+FN}
        \end{split}
    \end{equation}

    The metrics used in the evaluation procedure were the precision, recall and F1-Score. The latter being the weighted average of precision and recall, hence considering both false positives and false negatives into account to evaluate the performance of the model. 
    
    \begin{equation}
        F1-Score = 2\times \frac{(Recall \times Precision)}{(Recall + Precision)}
    \end{equation}

    \par
    Since we used cross validation techniques to evaluate the performance of each model, we calculated the mean ($\mu$) and standard deviation ($\sigma$) of the F1-scores of the model in order to have an average overview of its performance. The equations used are presented below: 

    \begin{equation}\label{mean and SD F1-score}
        \begin{split}
            \mu_{F1-Score} = \frac{\sum_{i=1}^{\mathcal{N}} (F1-Score_i)}{\mathcal{N}}
            \\
            \sigma_{F1-Score} = \sqrt{\frac{\sum_{i=1}^{\mathcal{N}}(F1-Score_i - \mu_{F1-Score})^2}{\mathcal{N}}}
        \end{split}
    \end{equation}
    
    where $\mathcal{N}$ is the number of splits generated from the cross validation procedure. For instance, one leave out generates five splits ($\mathcal{N}=5$) using \autoref{combination} as shown in  \autoref{fig:Stratified K-Fold Cross Validation leave one out}.  
    
    As for the loss metrics, we used the cross-entropy loss function between the true classes and predicted classes. 

\section{Results and Discussion}
State-of-the-art CNN-based architectures, ResNet and EfficientNet were trained along the ViT-B16 and ViT-B32 in order to compare their performances on our custom dataset comprising of 5 classes (weeds, beet, off-type beet, parsley and spinach). All models have been trained using the five-folds cross validation leaving one out technique. The accuracies and losses of the models tend to be flat after the 30th epoch. The average F1-Scores and losses obtained with 3211 testing images (original and unprocessed images except for off-type beet images) are reported in \autoref{tab:Models comparison table} below. 





\begin{table}[H]
    \centering
    \begin{tabularx}{0.5\textwidth}{ 
  | >{\raggedright\arraybackslash}X
  | >{\centering\arraybackslash}X
  | >{\centering\arraybackslash}X | }
    \hline
    Model & $\mu_{F1-Score}$ & Loss \\ \hline
    \textbf{ViT B-16} & \textbf{0.998  $\pm 0.002$} & \textbf{0.656} \\ \hline
    \textbf{ViT B-32} & \textbf{0.996  $\pm 0.002$} & \textbf{0.672} \\ \hline
    EfficientNet B0 & 0.987  $\pm 0.005$ & 0.735 \\ \hline
    EfficientNet B1 & 0.989  $\pm 0.005$ & 0.720\\ \hline
    ResNet 50 & 0.995  $\pm 0.005$ & 0.716 \\ \hline
    \end{tabularx}
    \caption{Comparison table between state-of-the-art CNN-based models and vision transformer models on agricultural image classification. The F1-Score has been calculated using \autoref{mean and SD F1-score} with $\mathcal{N}=5$.}
    \label{tab:Models comparison table}
\end{table}

From these experimental results, we notice the outperformance of the ViT models compared to the CNN-based models with a best F1-Score of 99.8\% for the ViT B-16 model although the ViT B-32 model's performance is very close behind at 99.6\% with a minimum loss of 0.656 for the ViT B-16 model. The EfficientNet and ResNet models fall behind compared to ViT models but with high scores nevertheless, having been trained on a large dataset (12 844 training images). These experimental results confirm vision transformers high performance compared to current state-of-the-art models ResNet and EfficientNet as presented by \citeauthor{Dosovitskiy2020}. Although all network families obtain high accuracy and precision, the classification of crops and weed images using vision transformer yields the best prediction performance. 

\subsection{Influence of the training set size}
In the next stage we tried to answer the question of which network family yields the best performance with a smaller training dataset. We did so by carrying out a five-folds cross validation leaving k out, where k is a varying parameter from 1 to 4 while keeping the testing set to 3211 images to evaluate the performance of the models. 

\begin{table*}[t]
\centering
\setlength\extrarowheight{20pt}
\resizebox{\textwidth}{!}{\Large%
\begin{tabular}{p{50mm}|lll||lll||lll||lll}

{\multirow{2}{*}{\diagbox{Classes}{k-Folds}}} & \multicolumn{3}{c||}{k=1} & \multicolumn{3}{c||}{k=2} & \multicolumn{3}{c||}{k=3} & \multicolumn{3}{c}{\textbf{k=4}} \\ \cline{2-13}
\multicolumn{1}{c|}{} & Prec & Rec & $\mu_{F1-Score}$ & Prec & Rec & $\mu_{F1-Score}$ & Prec & Rec & $\mu_{F1-Score}$ & Prec & Rec & $\mu_{F1-Score}$ \\ \cline{1-1}
 
Weeds & 1.000 & 1.000 & 1.000 $\pm 0.000$  & 1.000 & 0.989 & 0.994 $\pm 0.002$  & 1.000 & 0.987 & 0.993 $\pm 0.002$  & 0.993 & 0.976 & 0.985 $\pm 0.003$ \\ 

Off-Type Beet & 0.992 & 1.000 & 0.996 $\pm 0.001$  & 0.988 & 1.00 & 0.994 $\pm 0.002$  & 0.985 & 0.999 & 0.992 $\pm 0.002$  & 0.974 & 0.992 & 0.983 $\pm 0.004$ \\ 

Beet  & 1.000 & 1.000 & 1.000 $\pm 0.000$  & 0.999 & 1.000 & 0.999 $\pm 0.001$  & 0.999 & 1.000 & 0.999 $\pm 0.001$  & 0.998 & 0.999 & 0.999 $\pm 0.001$  \\ 

Parsley & 1.000 & 1.000 & 1.000 $\pm 0.000$  & 1.000 & 1.000 & 1.000 $\pm 0.001$  & 1.000 & 1.000 & 1.000 $\pm 0.000$  & 0.999 & 1.000 & 0.999 $\pm 0.000$  \\ 

Spinach & 1.000 & 1.000 & 1.000 $\pm 0.000$  & 1.000 & 1.000 & 1.000 $\pm 0.000$  & 1.000 & 1.000 & 1.000 $\pm 0.000$  & 0.999 & 1.000 & 0.999 $\pm 0.001$  \\ 
\end{tabular}%
}

\caption{Comparison of classification reports generated from 5-Fold cross validation leaving k folds out (where k-folds stands for the number of validation folds) with $1 \leq k \leq 4$. k = 1 represents the most number of training images (12 844) and k = 4 represents the lowest number of training images (3 211). The average precision (Prec), recall (Rec) and F1-Score, obtained using \autoref{mean and SD F1-score} are reported for each class obtained with the ViT B-16 model.}
\label{tab:comparison table kFold CV}
\end{table*}


Varying the number of training images has a direct influence on the performance of the trained ViT model, as shown in \autoref{tab:comparison table kFold CV}.  The results obtained with the five-folds cross validation, leaving two out as validation set (k=2) are promising, with a mean F1-Score of 99.7\% and a standard deviation of 0.1\% showing a very small difference between the scores of the 10 generated models. We notice a very small decrease in performance of the ViT B-16 model while reducing the number of training images. We note a very light decrease of 0.1\% in the accuracy of the ViT B-16 model while training only with 2/5 of the dataset (6422 images, k=3) and validating on the remaining 3/5. With k=4, the ViT B-16 model was trained with a smaller dataset of 3211 images (75\% reduction), and its performance decreased as expected but by a small margin of only 0.44\% for an overall accuracy of 99.63\%. These experimental results shows how well the vision transformer models perform with small datasets. The ViT B-16 model makes use of the self-attention mechanism to learn recognisable patterns from few images in order to achieve such high accuracy.

\begin{figure}[H]
    \centering
    \includegraphics[width=0.5\textwidth]{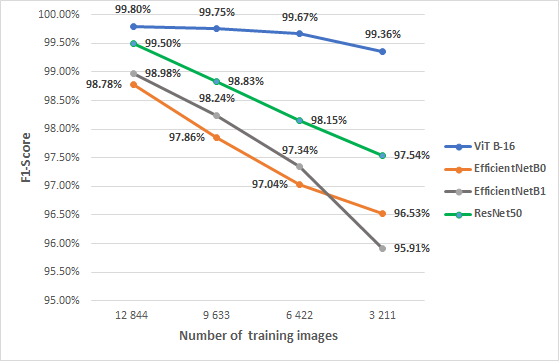}
    \caption{Comparison between ViT B-16, EfficientNet B0, EfficientNet B1 and ResNet50 on their respective performance with different number of training images.}
    \label{fig:Small Dataset Model Comparison}
\end{figure}

We then compared the performance of the ViT B-16 model to state-of-the-art CNN-based models ResNet and EfficientNet with a decreasing number of training images. The experimental results of their F1-Scores are reported in the \autoref{fig:Small Dataset Model Comparison}. We notice a decrease in the F1-Scores of the ResNet50, EfficientNet B0 and EfficientNet B1 with a reduction in the number of training images. In contrast, the ViT B-16 model keeps its high performance in this set of experiments, specially with the smallest number of training images, achieving an F1-Score of 99.63\%. On the other hand, RestNet50 scores an accuracy of 97.54\%, EfficientNet B0 scores 96.53\% and EfficientNet B1 with the worst score of 95.91\%. EfficientNet B1 has the worst decrease in performance of 3.07\% (from 98.98\% - with 12 844 training images to 95.91\% - with 3211 training images). Even though EfficientNet B1 achieves better results with the largest dataset (98.98\% accuracy) than EfficientNet B0 (98.78\%), its performance falls off the most with the smallest training dataset. While the F1-Scores of ResNet and EfficientNet B0 and B1 declines with a reduction of training images by 25\% (from 12 844 images to 9 633 images), the ViT B-16 model still achieves a high performance of 99.75\% (a slight decrease from 99.80\%). These experimental results show the outperformance of vision transformer models over current CNN-based models ResNet and EfficientNet in agricultural image classification when dealing with small training datasets.

\begin{table}[H]
\centering
\resizebox{0.45\textwidth}{!}{%
\begin{tabular}{l|llll}
\hline
\\
Number of testing images & ViT B-16 & EfficientNetB0 & EfficientNetB1 & ResNet50 \\ \hline
\\
3211 & 99.80\% & 98.78\% & 98.98\% & 99.50\% \\ \hline
\\
9633 & 99.61\% & 96.60\% & 98.59\% & 98.59\% \\ \hline
\\
13596 & 99.14\% & 97.40\% & 94.10\% & 95.32\% \\ \hline
\end{tabular}%
}
\caption{Comparison of performances of ViT B-16, EfficientNet B0, EfficientNet B1 and ResNet50 with a decreasing number of training images while using a 5-Fold cross validation technique leaving one fold out as validation. The average F1-Scores for each model, obtained using \autoref{mean and SD F1-score} are reported for three set of experiments with a varying number of testing images.}
\label{comparison table testing sets}
\end{table}

\begin{figure}[]
    \centering
    \includegraphics[width=0.5\textwidth]{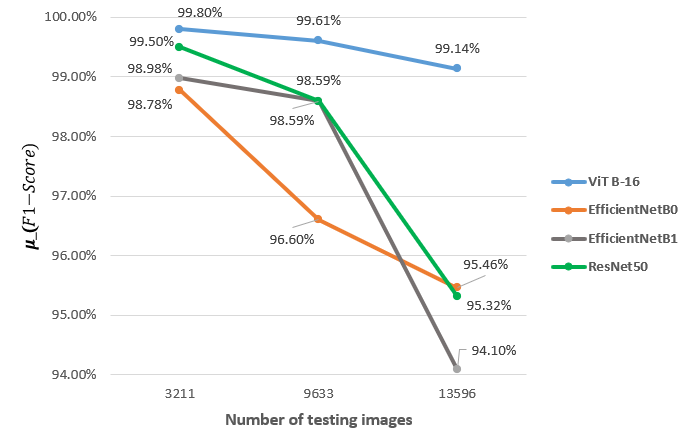}
    \caption{Comparison between ViT B-16, EfficientNet B0, EfficientNet B1 and ResNet50 on their respective performance with different number of testing set while keeping 5-Fold cross validation leaving one fold as validation set.}
    \label{fig:Model comparison testing set}
\end{figure}

Furthermore, we compared the performance of the models with a variation in the number of testing images while using a 5-Folds leaving one fold out cross validation technique. The F1-Scores are reported in \autoref{comparison table testing sets}. As shown in \autoref{fig:Model comparison testing set}, 
there is a notable decrease in the F1-Scores of the four models while testing with 9633 and 13596 images and training with only 50\% and 30\% of the labelled dataset. On the third set of experiments, the models were trained on only 4535 images and validating on 1134 images, which explains the decrease in their performances. Even though all models have a decrease in their F1-Scores with an increasing number of testing images, the ViT B-16 model still achieves higher performance than the state-of-the art CNN-based models, EfficientNetB0, EfficientNetB1 and ResNet50. The ViT B-16 model had the smallest decrease in performance from 99.80\% (for 3211 testing images and 12 844 training images) to 99.14\% (for 13 596 testing images and 4535 training images).

\section{Conclusion}
In this study, we used the self-attention paradigm via the vision transformer models to learn and classify custom crops and weeds images acquired by UAV in beet, parsley and spinach fields. The results achieved with these datasets indicate a promising direction in the use of vision transformers in agricultural problems. Outperforming current state-of-the-art CNN-based models like ResNet and EfficientNet, the base ViT model is to be preferred over the other models for its high accuracy and its low calculation cost. Furthermore, the ViT B-16 model has proven better with its high performance specially with small training datasets where other models failed to achieve such high accuracy. This shows how well the convolutional-free, ViT model interprets an image as a sequence of patches and processes it by a standard transformer encoder, using the self-attention mechanism, to learn patterns between weeds and crops images. In this respect, we come to conclusion that the application of vision transformer could change the way to tackle vision tasks in agricultural applications for image classification by bypassing classic CNN-based models. In future works, we plan to use vision transformer classifier as a backbone in an object detection architecture to locate and identify weeds and plants on UAV orthophotos.

\section*{Acknowledgement}
This work was carried out as a part of DESHERBROB (2020-2024) project funded by the Region Centre-Val de Loire (France). We gratefully acknowledge Region Centre-Val de Loire for its support. We thank the company FRASEM, partner of this project for their valuable provision of plots and data.

\bibliographystyle{IEEEtranN}
\bibliography{reference}

\end{document}